# Improved AdaBoost for virtual reality experience prediction based on Long Short-Term Memory network

**Wenhan Fan[1,6,*], Zhicheng Ding[2], Ruixin Huang[3], Chang Zhou[4], Xuyang Zhang[5]**

[1]Independent Researcher, New York , NY, 10019, USA.
[2]Fu Foundation School of Engineering and Applied Science, Columbia University, New York, NY, 10027, USA.
[3]Carnegie Mellon University, ECE, Pittsburgh, PA, 15206, USA.
[4]Independent Researcher, Kirkland, WA, 98034, USA.
[5]Independent researcher, New York, 11101, USA.

[6]finncontactplus@gmail.com
*corresponding author

**Abstract.** A classification prediction algorithm based on Long Short-Term Memory Network (LSTM) improved AdaBoost is used to predict virtual reality (VR) user experience. The dataset is randomly divided into training and test sets in the ratio of 7:3.During the training process, the model's loss value decreases from 0.65 to 0.31, which shows that the model gradually reduces the discrepancy between the prediction results and the actual labels, and improves the accuracy and generalisation ability. The final loss value of 0.31 indicates that the model fits the training data well, and is able to make predictions and classifications more accurately. The confusion matrix for the training set shows a total of 177 correct predictions and 52 incorrect predictions, with an accuracy of 77%, precision of 88%, recall of 77% and f1 score of 82%. The confusion matrix for the test set shows a total of 167 correct and 53 incorrect predictions with 75% accuracy, 87% precision, 57% recall and 69% f1 score. In summary, the classification prediction algorithm based on LSTM with improved AdaBoost shows good prediction ability for virtual reality user experience. This study is of great significance to enhance the application of virtual reality technology in user experience. By combining LSTM and AdaBoost algorithms, significant progress has been made in user experience prediction, which not only improves the accuracy and generalisation ability of the model, but also provides useful insights for related research in the field of virtual reality. This approach can help developers better understand user requirements, optimise virtual reality product design, and enhance user satisfaction, promoting the wide application of virtual reality technology in various fields.

**Keywords:** LSTM, AdaBoost, Virtual Reality.

## 1. Introduction

Virtual Reality (VR) technology, as an emerging interactive experience, has been widely used in entertainment, education, and medical fields. User experience research aims to understand users' feelings, attitudes and behaviours when using VR products or services in order to guide product design and improvement [1]. Contextually, the rapid development and popularity of virtual reality technology has led to an increasing emphasis on user experience. Traditional user experience research methods often rely on user feedback surveys and observations, but these methods have problems such as high subjectivity and limitations. Therefore, combining machine learning algorithms to predict user experience has become one of the hot spots in current research.

Machine learning algorithms play an important role in predicting user experience. First, by collecting a large amount of user data and behavioural information, machine learning algorithms can help analyse the laws and patterns hidden behind the data, thus revealing the key factors affecting user experience [2,3]. Secondly, machine learning algorithms can build predictive models to predict the performance of user experience under new users or new scenarios in the future by training and learning from existing data. This data-driven based approach not only improves prediction accuracy, but also enables personalised user experience prediction.

In addition, machine learning algorithms can help identify the differences between different user groups, thus providing a basis for product customisation. Through technical means such as cluster analysis, users can be divided into different groups, and corresponding user experience strategies can be designed for different groups [4]. This kind of refined personalised design can better meet the needs and preferences of different user groups and improve overall user satisfaction.

In conclusion, as an interdisciplinary field, virtual reality user experience research shows greater potential after integrating machine learning algorithms. With the continuous development and improvement of technology, machine learning algorithms will play a more and more important role in the field of virtual reality in the future, bringing more high-quality, personalised user experience.

## 2. Data set sources and data analysis

The dataset used in this paper is selected from the Kaggle public dataset, which consists of user experiences in a virtual reality (VR) environment. It includes data related to physiological responses such as heart rate and skin conductance, emotional states and user preferences. The aim of this dataset is to enhance VR technology by analysing user experience. The analysis aims to improve VR design, user comfort and customisation by understanding users' physiological and emotional responses to different VR environments [5]. The dataset enables developers to optimise VR systems and create tailored experiences to enhance immersion and overall user satisfaction. Some of the data is shown in Table 1.

**Table 1.** Partial text data.

| Age | Gender | VRHeadset | Duration | MotionSickness | ImmersionLevel |
|---|---|---|---|---|---|
| 40 | Male | HTC Vive | 13.59850823 | 8 | 5 |
| 43 | Female | HTC Vive | 19.95081498 | 2 | 2 |
| 27 | Male | PlayStation VR | 16.5433874 | 4 | 2 |
| 46 | Other | PlayStation VR | 48.88756499 | 6 | 2 |
| 54 | Male | HTC Vive | 53.91962373 | 5 | 2 |
| 38 | Female | Oculus Rift | 18.42890716 | 7 | 1 |
| 35 | Other | PlayStation VR | 44.10541934 | 5 | 3 |

## 3. Method

### 3.1. Long Short-Term Memory (LSTM) network

Long Short-Term Memory (LSTM) network is a deep learning model commonly used to process sequential data, especially suitable for tasks requiring long-term memory and capturing long-distance dependencies in a time series. Inspired by the workings of neurons in the human brain that are responsible for processing memories and learning, the LSTM network achieves long-term storage and selective forgetting of information in a sequence of data through a refined structure [ 6,7].

LSTM networks consist of a series of structures called "gates", including forgetting gates, input gates and output gates. These gates control the flow of information through the network, thus enabling the processing of information from different parts of the data sequence. The schematic diagram of LSTM network is shown in Figure 1.

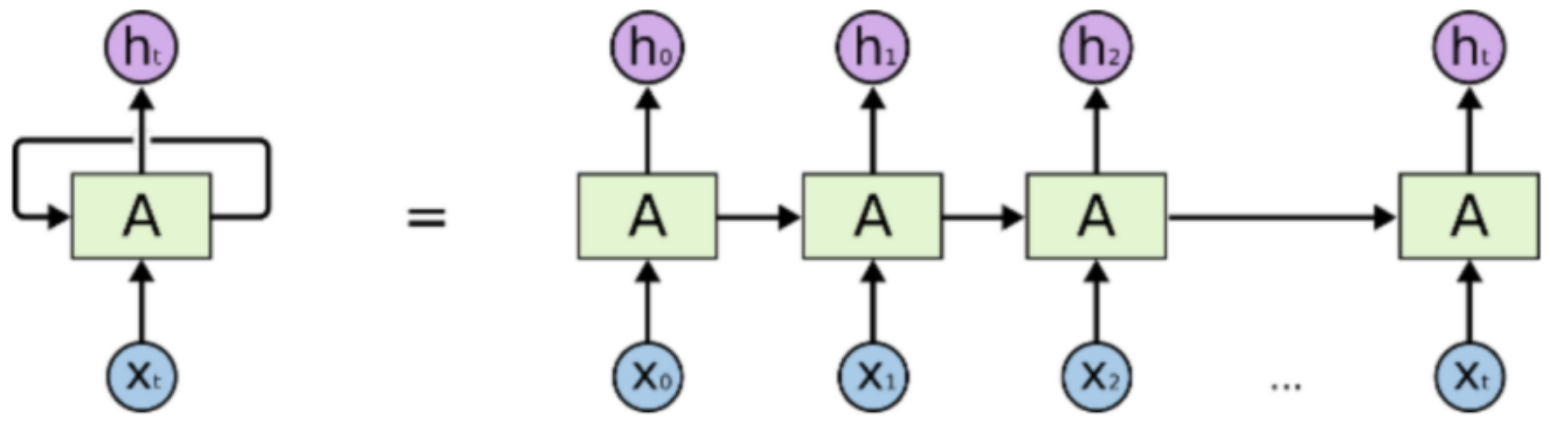

**Figure 1.** The schematic diagram of LSTM network.
（Photo credit: Original）

The input gate determines which information can enter the cell state. It consists of a Sigmoid layer, which outputs values ranging from 0 to 1, indicating what percentage of each input message can pass, and a Tanh layer, which generates a new vector of candidate values. The Oblivion Gate determines which information should be discarded from the cell state. It consists of a Sigmoid layer with output values also ranging from 0 to 1, indicating what proportion of the information in the corresponding position in each cell state should be retained. The cell states are updated based on input gates, forgetting gates and a vector of candidate values. Firstly, the forgetting gate is used to decide which information to discard, and then the input gate is used to control the new information entering the cell state. In this process, the previous moment's value in the storage cell is forgotten and only useful information is retained. As shown in equation (1):

$$ft=\sigma(Wf1ht-1+W02Xt+bf) \quad (1)$$

The output gate determines the hidden state of the final output. It consists of a Sigmoid layer and a Tanh layer.The Sigmoid layer determines which part of the unit state will be output, and the Tanh layer compresses this unit state value between -1 and 1 [8].

With the above structure, the LSTM network is able to process sequential data efficiently, and through a well-designed memory mechanism, the LSTM is able to maintain long-term dependencies. The forgetting of unimportant information from the past is controlled through forgetting gates to maintain network efficiency. The information to be processed at the current moment is controlled through input and output gates.The LSTM network has a trainable set of parameters that can automatically adjust the weights according to the task.

### *3.2. AdaBoost*

AdaBoost (Adaptive Boosting) is an integrated learning method designed to improve the performance of weak classifiers by combining multiple weak classifiers to build a strong classifier.The core idea of the AdaBoost algorithm is to iteratively train a series of basic classifiers and adjust the weights of the samples according to the performance of each basic classifier, so that the previously misclassified samples that were previously misclassified receive more attention in the next round of training, thus continuously improving the accuracy of the overall model [9]. In the derivation of the AdaBoost algorithm, the more easily understood logic is the "additive model", i.e., for a linear combination of learners, as shown in equations (2) and (3):

$$H(x)=\sum Tt=1atht(x) \quad (2)$$

$$l(h|d)=Ex\sim d[e-f(x)h(x)] \quad (3)$$

At the beginning, equal weights are given to each sample, i.e., each sample contributes equally to the final classifier. In each iteration, a base classifier (usually a weak classifier) is selected and it is trained with the current sample weights. After training is complete, the error rate of this classifier on the training set is calculated. The sample weights are adjusted according to the performance of the current classifier on the training set. Misclassified samples will be given higher weights while correctly classified samples will be given less weights. The weights of the current base classifiers are calculated based on their error rates on the training set and are added to the final combined model [10]. It also affects the weights corresponding to each sample in the next iteration. When a predetermined number of iterations is reached or a certain performance metric is achieved, the iterations are stopped and the final combined model is output.

By constantly adjusting the sample weights and the basic classifier weights, the AdaBoost algorithm focuses on those previously misclassified samples in each iteration, gradually improving the overall model performance in the process of continuous iteration. Due to its excellent generalisation ability and strong robustness to noisy data, it is widely used in practical applications to solve binary classification, multi-classification and other problems.

The classification algorithm based on Long Short-Term Memory Network (LSTM) to improve AdaBoost uses LSTM as the base classifier, and takes advantage of its good modelling ability on time series data to capture the temporal information in the data. In each round of AdaBoost iteration, the model performance is continuously improved by training the LSTM base classifier and adjusting the sample weights according to its prediction results, so that the next round of training pays more attention to previously misclassified samples. By combining the sequence modelling capability of LSTM and the focus on misclassified samples in the AdaBoost integrated learning framework, the model can effectively handle time series data, overcome the overfitting problem, and gradually improve the overall classifier performance in each iteration, thus achieving more accurate classification predictions.

## 4. Result

In this experiment, Matlab R2022a is used for training, and the training and test sets are randomly divided according to the ratio of 7:3, with 70% of the data used for training and 30% for testing. The maximum number of training times is set to 50, the initial learning rate is 0.01, the learning rate decrease factor is set to 0.1, the loss change curve during training is recorded, as shown in Fig. 2, and the prediction confusion matrices of the training set and validation set are recorded at the same time, as shown in Fig. 3 and Fig. 4, and the evaluation indexes of the training set and test set of the model are shown in Table 2.

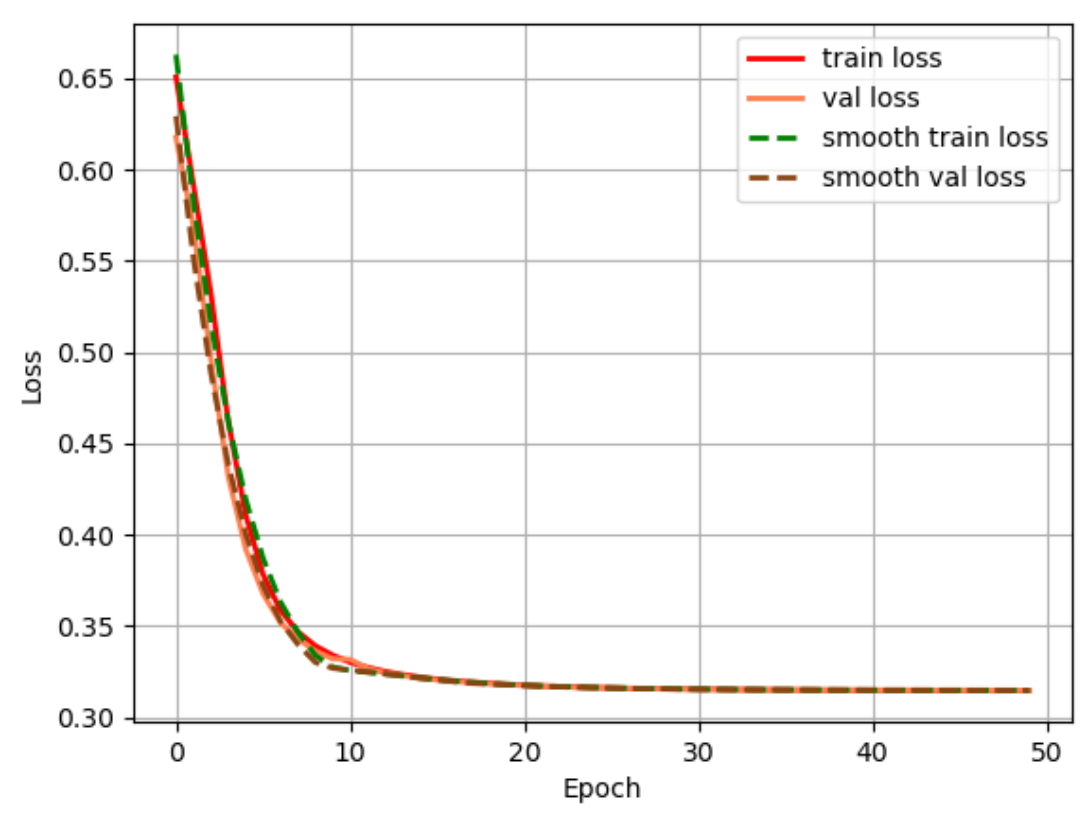

**Figure 2.** Loss.
（Photo credit: Original）

During the training process, the loss value decreases from 0.65 to 0.31, and the model gradually reduces the discrepancy between the prediction results and the actual labels during the learning process, which improves the accuracy and generalisation of the model. The final loss value of 0.31 means that the model fits the training data well and can predict and classify more accurately.

**Table 3.** Model evaluation parameter.

| Evaluation parameters | Train | Test |
|---|---|---|
| Accuracy | 0.77 | 0.75 |
| Precision | 0.88 | 0.87 |
| Recall | 0.77 | 0.57 |
| F1 Score | 0.82 | 0.69 |

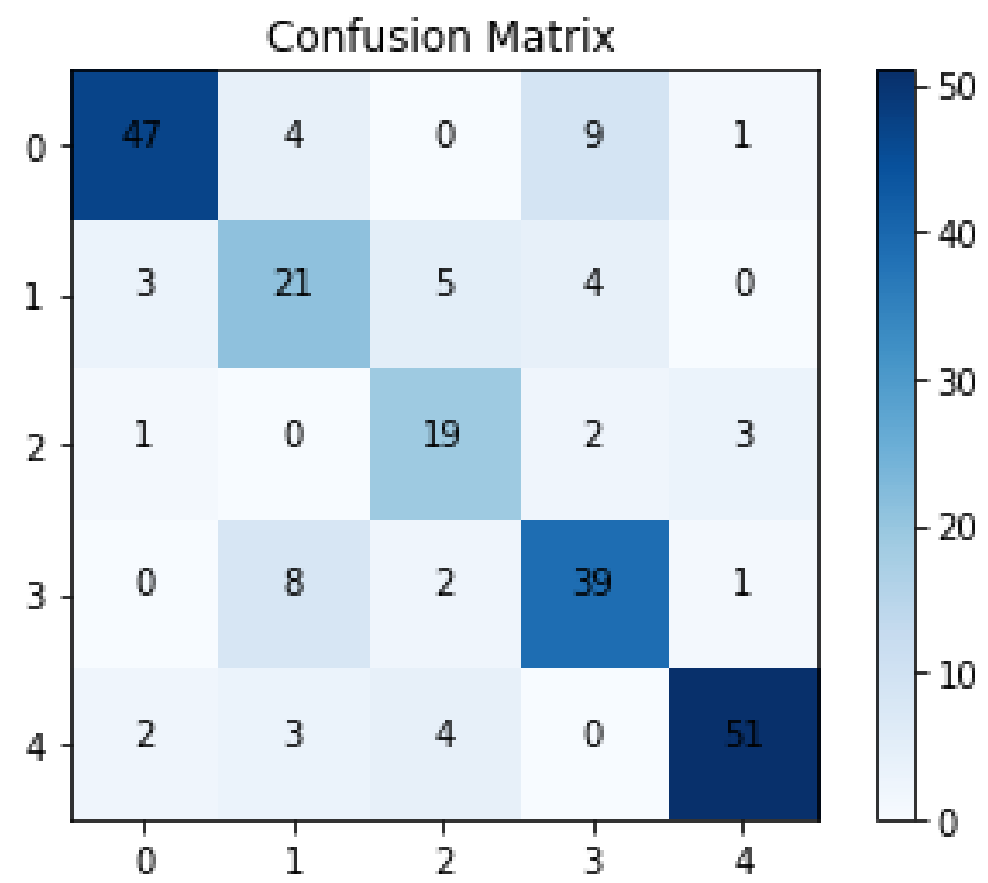

**Figure 2.** Confusion matrix.
（Photo credit: Original）

From the confusion matrix of the training set, there are 177 correct predictions and 52 incorrect predictions, the accuracy of the training set is 77%, the collinearity is 88%, the recall is 77% and the f1 score is 82%.

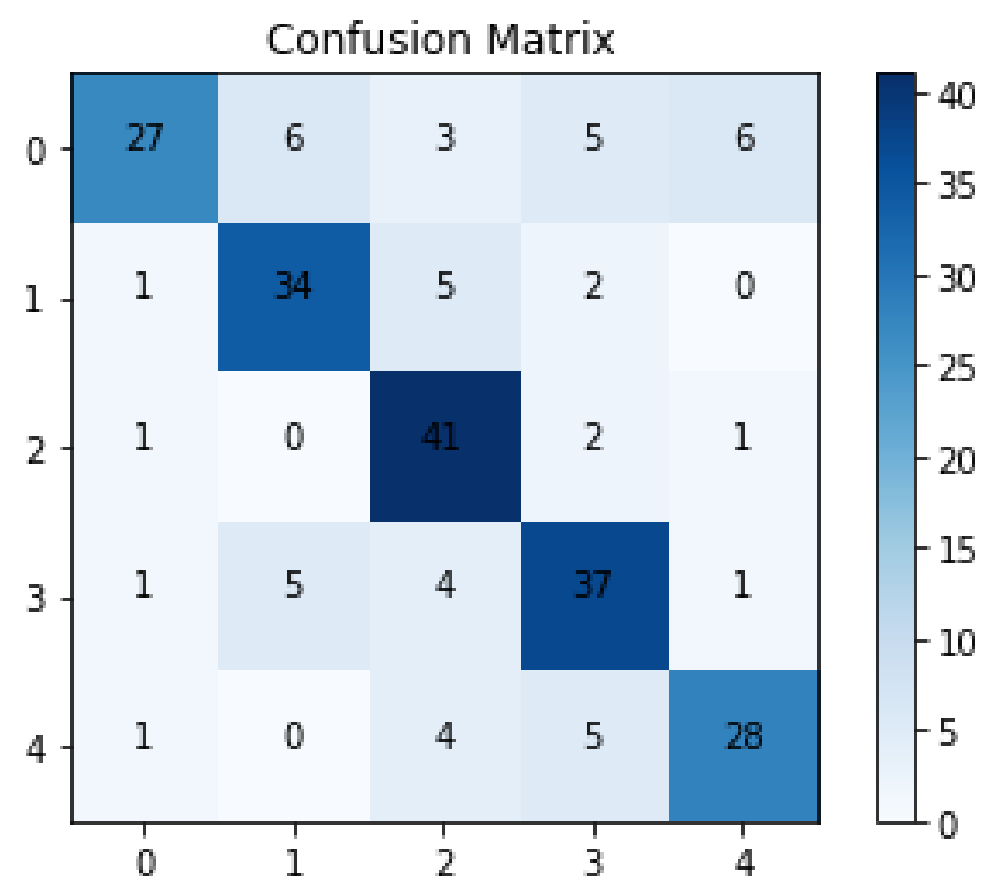

**Figure 3.** Confusion matrix.
（Photo credit: Original）

From the confusion matrix of the training set, there are 167 correct predictions and 53 incorrect predictions, and the accuracy of the training set is 75%, the collinearity is 87%, the recall is 57%, and the f1 score is 69%.

In summary, the classification prediction algorithm based on Long Short-Term Memory Network (LSTM) with improved AdaBoost is able to predict virtual reality users' experience well.

## 5. Conclusion

In this study, a classification prediction algorithm based on Long Short-Term Memory Network (LSTM) with improved AdaBoost was used with the aim of making accurate predictions of virtual reality (VR) user experience. The dataset was randomly divided into a training set and a test set in the ratio of 7:3. During the training process, the loss value of the model decreases from 0.65 to 0.31, indicating that the model gradually reduces the discrepancy between the prediction results and the actual labels, and improves the accuracy and generalisation ability. The final loss value of 0.31 indicates that the model fits the training data well and is able to predict and classify more accurately. According to the training set confusion matrix, there were 177 correct predictions and 52 incorrect predictions with 77% accuracy, 88% precision, 77% recall and 82% F1 score. Whereas, the test set confusion matrix shows that there are 167 correct and 53 incorrect predictions with 75% accuracy, 87% precision, 57% recall and 69% F1 score.

In summary, this study improved the AdaBoost algorithm based on LSTM successfully applied to the prediction of virtual reality user experience and achieved better results. Through this algorithm, the accuracy of user experience prediction is improved while the generalisation ability of the model is also enhanced, which provides effective support and guidance for the development of virtual reality technology. In the future, the algorithm parameters can be further optimised, the data scale can be expanded and more feature engineering methods can be explored to further improve the model performance and application scope.